\documentclass{article}

\usepackage{PRIMEarxiv}

\usepackage[utf8]{inputenc} 
\usepackage[T1]{fontenc}    
\usepackage{url}            
\usepackage{booktabs}       
\usepackage{amsfonts}       
\usepackage{nicefrac}       
\usepackage{microtype}      
\usepackage{lipsum}
\usepackage{fancyhdr}       
\usepackage{graphicx}       
\usepackage{color}
\usepackage{ulem}

\usepackage{threeparttable}
\usepackage{colortbl}
\usepackage[table]{xcolor}
\usepackage{threeparttable}
\usepackage{tabularx}
\usepackage{kotex}
\usepackage{xcolor}
\usepackage{times}
\usepackage{epsfig}
\usepackage{graphicx}
\usepackage{amsmath}
\usepackage{amssymb}
\usepackage{multirow}
\usepackage{subfigure}
\usepackage{float}
\usepackage{url}
\usepackage[colorlinks=true, linkcolor=blue, citecolor=blue, urlcolor=blue]{hyperref}
\definecolor{lightgray}{gray}{0.9}
\definecolor{pastelyellow}{RGB}{255, 255, 230}
\definecolor{pastelred}{RGB}{255, 230, 230}
\definecolor{mint}{RGB}{0, 255, 255}

\graphicspath{{media/}}     

\title{ 
    Flatfish Lesion Detection 
    \\ Based on Part Segmentation Approach 
    \\ and Lesion Image Generation
}

\author{
  Seo-Bin Hwang \\
  Dept. AI Convergence\\
  Chonnam National University \\
  Gwang-ju\\
  \texttt{cnu.cvl.hsb@gmail.com} \\
  \And
  Han-Young Kim \\
  Dept. AI Convergence\\
  Chonnam National University \\
  Gwang-ju\\
  \texttt{codebyhy@gmail.com} \\
   \And
   Chae-Yeon Heo \\
  Dept. AI Convergence\\
  Chonnam National University \\
  Gwang-ju\\
  \texttt{214274@jnu.ac.kr} \\
   \And
   Hie-Yong Jeong \\
  Dept. AI Convergence\\
  Chonnam National University \\
  Gwang-ju\\
  \texttt{h.jeong@jnu.ac.kr} \\
   \And
   Sung-Ju Jung \\
  Dept. Marine and Fisheries Science\\
  Chonnam National University \\
  Gwang-ju\\
  \texttt{sungju@chonnam.ac.kr} \\
   \And
   Yeong-Jun Cho\thanks{Corresponding author}\\
  Dept. AI Convergence\\
  Chonnam National University \\
  Gwang-ju\\
  \texttt{yj.cho@jnu.ac.kr} \\
}

\begin{document}
\maketitle
\begin{abstract}
The flatfish is a major farmed species consumed globally in large quantities. 
However, due to the densely populated farming environment, flatfish are susceptible to injuries and diseases, making early disease detection crucial. 
Traditionally, diseases were detected through visual inspection, but observing large numbers of fish is challenging. 
Automated approaches based on deep learning technologies have been widely used, to address this problem, but accurate detection remains difficult due to the diversity of the fish and the lack of the fish disease dataset.
In this study, augments fish disease images using generative adversarial networks and image harmonization methods. 
Next, disease detectors are trained separately for three body parts (head, fins, and body) to address individual diseases properly.
In addition, a flatfish disease image dataset called \texttt{FlatIMG} is created and verified on the dataset using the proposed methods.
A flash salmon disease dataset is also tested to validate the generalizability of the proposed methods.
The results achieved 12\% higher performance than the baseline framework.
This study is the first attempt to create a large-scale flatfish disease image dataset and propose an effective disease detection framework.
Automatic disease monitoring could be achieved in farming environments based on the proposed methods and dataset. 

\end{abstract}


\section{Introduction}
	\label{sec:introduction}
	
	Flatfish have been highly farmed all over the world and consumed worldwide, especially in Japan and South Korea~\cite{huang2015novo,url:Comprehensive-Fisheries-Information-System}.
	They are referred to as demersal fish, also known as groundfish, which lack air bladders and usually inhabit the seabed.
	Due to the ecology of flatfish and the densely populated aquaculture environment, 
	flatfish are likely to overlap and bump into each other on the fishery grounds.
	Such farming conditions make flatfish susceptible to injury and infection.
	Therefore, early disease detection is critical to prevent the spread of infections and ensure increased productivity in flatfish aquaculture. 
	Generally, experienced aquaculturists visually check fish to detect diseases~\cite{hasan2022fish} and perform additional inspections, such as a biopsy for fish displaying disease symptoms.
	However, visually inspecting fish is challenging.
	First, the number of fish is too high for a limited number of aquaculturists to investigate them.
	Second, the fish must be pulled from the fishery for inspection, which is stressful and inconvenient for both the workers and the fish. Therefore, many fish disease detection methods~\cite{kumar2017fish, al2022hybrid, hasan2022fish} that automatically detect disease areas from fish images have been proposed to address these challenges.
 
	Recently, several studies~\cite{hasan2022fish, al2022hybrid} have proposed fish disease detection methods based on convolutional neural networks~(CNNs).  
        Although many studies have been proposed to detect fish diseases, no studies have been conducted on flatfish disease detection using a large-scale flatfish dataset.  

	For example,  the hybrid CNN~\cite{al2022hybrid} has been used to test sufficient diseased fish images, but the target fish were not flatfish, but rohu fish. 
	Some studies~\cite{kumar2017fish, malik2017image, hasan2022fish} have addressed disease detection but did not explain the details of the image datasets or only validated the methods on a small dataset.
	
	\begin{figure*}
		\centering
		\subfigure[Diseases on head]{\includegraphics[width=0.3\linewidth]{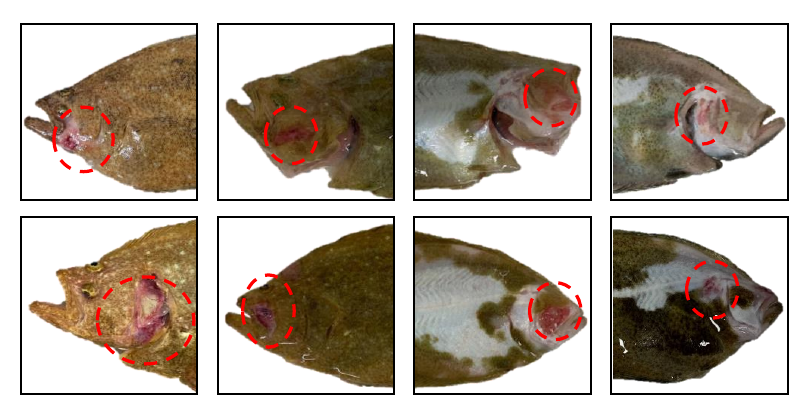}} \hspace{10pt}
		\subfigure[Diseases on body]{\includegraphics[width=0.3\linewidth]{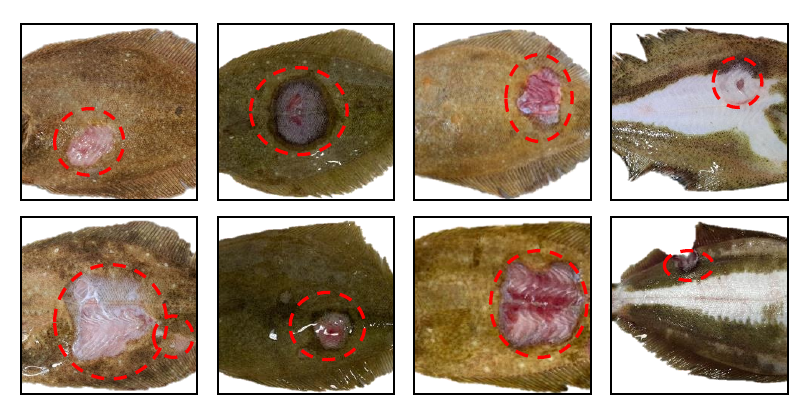}} \hspace{10pt}
		\subfigure[Diseases on fins]{\includegraphics[width=0.3\linewidth]{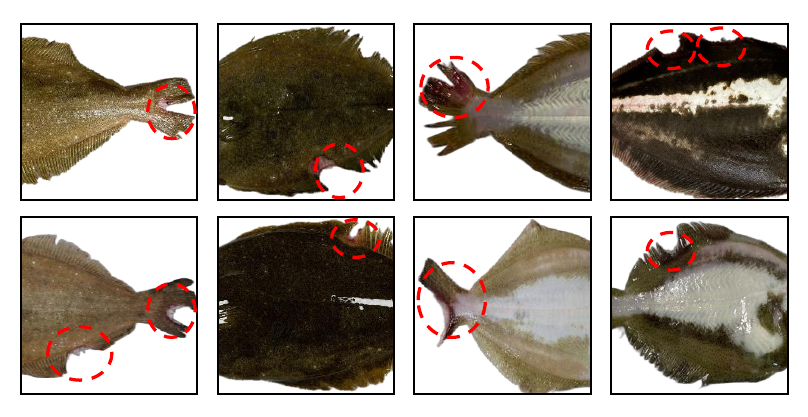}}
		\caption{Disease appearance by flatfish body part. (a) Flatfish diseases on the eyes, mouth, and gills. (b) Flatfish diseases on the back and stomach. (c) Flatfish diseases on the upper, lower, and tail fins.}
		\label{fig:appearance-bodyparts}
	\end{figure*}
	
	This study contributes to generating a flatfish disease image dataset and proposes an effective disease detection framework to overcome the limitations of previous studies.  
	For the disease image datasets, we have collected new diseased flatfish images called \texttt{flatIMG} including $164$ numbers of fish identities from the $394$ disease identities.
	It includes the various pathogens such as bacteria, viruses, and parasites occurring in fish diseases as summarized in Table.~\ref{tab:typesofdis}. 
	Although \texttt{flatIMG} is the largest dataset for flatfish diseases, in Section~\ref{sub:dataset_gen}, we propose diseased fish image-generation methods to augment its volume.
	To this end, we train a fish disease patch generation model based on generative adversarial networks~(GAN)~\cite{goodfellow2014generative}.
	We further mix the generated disease patches onto non-diseased flatfish images through the proposed image harmonization framework.
	As a result, we finally create $4,773$ diseased fish image datasets.
 
	After collecting and augmenting large-scale flatfish image datasets, we observed that the symptoms of diseases can be displayed anywhere on the flatfish's body.
	For example, scuticociliatosis abnormally changes the skins and organs of fish as depicted in Fig~\ref{fig:appearance-bodyparts}.
	The symptoms were exposed on the eyes, mouth, gill, body, and fins. 
	In addition, it was confirmed that the appearance of disease symptoms is more related to the body parts of the fish rather than the types of diseases and pathogens.
 	In other words, the appearance of diseases can be categorized depending on where the disease occurred.
	The goal of this study is not to classify disease or pathogen types, but to detect diseases immediately whether any symptoms have been visually exposed to the fish.
	
	To achieve this, we focus only on the visual features of diseases.
	Therefore, we divide the flatfish into three different body parts (i.e., head, body, and fins) and handle them separately.
	To divide the flatfish, we propose effective three-part segmentation methods in Section~\ref{sec:Pb_fish_detector}.
	Before the part segmentation, all flatfish images are horizontally aligned through the proposed steps. 
	Then, several critical points are estimated from the aligned flatfish images to determine the three parts: head, body, and fins as shown in FIGURE~\ref{fishpartsegment}.
	The proposed segmentation methods do not require additional annotations of the pixel labels or burdensome deep learning algorithms~\cite{yu2018bisenet}. 
	After dividing the parts of the flatfish, a flatfish disease detector is trained separately according to the part with the disease. 
	The separate training of disease detectors for each dividing part resulted in a 7\% improvement in performance compared to detecting them as a single disease class.
	Furthermore, by utilizing the proposed image-generation method for data augmentation, there was a 12\% improvement in performance compared to its baseline framework.
        
	In summary, the contributions of this study are as follows:
	\begin{itemize}
		\item[$\bullet$] 
		Building a new flatfish image dataset containing various diseases,
		\item[$\bullet$] 
		Augmenting images of diseased fish based on a generative model and image harmonization,
		\item[$\bullet$] 
		Proposing an effective flatfish disease detector via part segmentation.
	\end{itemize}
	To the best of our knowledge, this is the first attempt to build the dataset of flatfish and propose an effective disease detector for flatfish.
	We hope that our study will provide useful guidelines to readers who desire to implement flatfish detectors.
	
	\begin{figure*}
		\centering
		\includegraphics[width=1\linewidth]{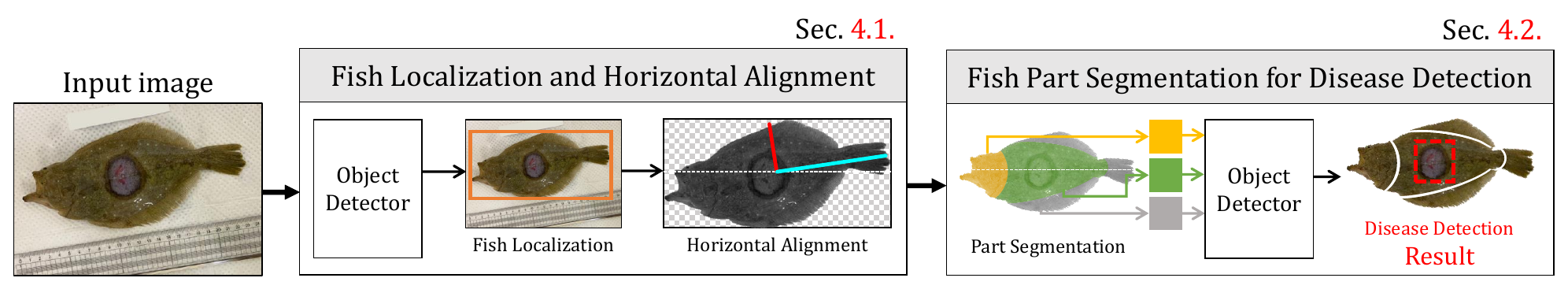}
		\caption{The overall framework for flatfish disease detection. First, it automatically detects a flatfish in the image and horizontally aligns the fish in Section~\ref{subsec:local_alignment}. Then it further performs three parts segmentation in Section.~\ref{sec:Three-part split}. Finally, it separately handles the three parts and detects the diseases of flatfish.}
		\label{fig:framework}
	\end{figure*}
	\section{Related Works}
	\label{sec:Related works}
	
	\subsection{Flatfish Disease Image Datasets and Image Augmentation}
	
	To the best of our knowledge, there is no public flatfish image dataset including various diseases.
	An image flatfish dataset~\cite{url:ai-hub} exists but only provides non-diseased flatfish images with their location and size annotations.
	Some studies~\cite{kumar2017fish, al2022hybrid, hasan2022fish} have proposed fish disease methods but did not offer fish disease datasets.
	However, some datasets~\cite{Sikder2021,ulucan2020large,fish-skin-diseases-classification_dataset} provide diseased images of fish, such as catfish, sea bass, or goldfish, but do not include flatfish. 
	Furthermore, the volumes of these public datasets of fish diseases are relatively small compared to other common image datasets, such as the ImageNet~\cite{deng2009imagenet} and MSCOCO~\cite{lin2014microsoft}.
	
	Many image augmentation methods have been proposed based on the given datasets to overcome the lack of image datasets.
	Traditionally, geometric transformations~\cite{maharana2022revieww}, photometric adjustments, and noise insertions are commonly utilized for augmentation.
	However, these simple color and geometric transformations highly rely on the original distribution of the reference images.
	Hence, the variations of the augmented images are very limited.
	
	Recently, generative models using deep neural network structures have been utilized for augmenting the dataset.
	Especially, several specific research fields such as agriculture, medicine, and aquaculture, which are difficult to collect sufficient datasets, have adopted the generation models.
	For example, Abbas~\textit{et al.}~\cite{abbas2021tomato} augmented specific conditions of diseased tomato leaves based on cconditional GANs\cite{mirza2014conditional}. 
	Similarly, Xiao~\textit{et al.}~\cite{xiao2022citrus} proposed a yellow dragon disease detection for citrus, stating that obtaining a sufficient dataset of leaves affected by yellow dragon disease was challenging because infected citrus plants are immediately removed. 
	To handle the lack of diseased citrus images, they combined a region-based CNN(R-CNN)~\cite{hee2017mask} with the CycleGAN~\cite{zhu2017unpaired} and built a style transfer to generate diseased leaves from healthy leaves.
	
	Image augmentation has been applied in the agricultural domain but not in aquaculture or fishery studies.
	Furthermore, although studies~\cite{abbas2021tomato, xiao2022citrus} have augmented images by transforming the overall style of leaves or fruits, diseases in animals, such as fish, have more complex patterns.
	Therefore, these image augmentation trials in previous works~\cite{abbas2021tomato, xiao2022citrus} cannot be used in flatfish disease generation.
	
	\subsection{Fish Disease Detection}
	
	Many studies have proposed fish disease detectors based on image processing and machine learning methods to detect fish diseases automatically.  
	For example, Kumar~\textit{et al.}~\cite{kumar2017fish} addressed epizootic ulcerative syndrome (EUS), which is a fish disease caused by the fungus \textit{Aphanomyces invadans}.
	They focused on detecting fish diseases by extracting FAST and histogram of oriented gradients (HOG) features~\cite{dalal2005histograms} from the fish images. In addition, Malik~\textit{et al.}\cite{malik2017image}. proposed a detector for EUS by employing various edge detectors for segmentation.
	However, the methods for fish disease detection following these conventional algorithms often fail to detect new disease patterns.
	With the recent development of deep learning, many attempts to adopt deep learning in fish disease detection have been presented.
	For example, one study~\cite{al2022hybrid} proposed a hybrid-CNN to detect diseases of the rohu fish by combining three CNN~\cite{krizhevsky2012imagenet} models such as VGG16~\cite{simonyan2014very}, Xception~\cite{chollet2017xception}, and DenseNet201~\cite{huang2017densely}. They tested $1,672$ diseased rohu fish and performed an accuracy of 99.82\% on their dataset.
	Another study~\cite{hasan2022fish} also proposed a method for detecting fish diseases using CNNs.
	It utilized a total of 60 diseased and 30 non-diseased fish images consisting of white and red spot diseases and performed 94.44\% accuracy on their dataset.
	
	Although many studies have been proposed to detect fish diseases, no studies have been conduted on flatfish disease detection using a large-scale flatfish dataset.
	Studies~\cite{kumar2017fish, malik2017image} have addressed EUS disease detection but did not provide details on the image datasets. Hasan \textit{et al.}~\cite{hasan2022fish} validated their methods on very small image datasets containing $60$ diseased fish.
	Compared to the previous studies, the present study addresses a new large-scale flatfish image dataset for the public and provides the disease detection method for flatfish, considering their characteristics.
	
	\section{Part Segmentation Approach for Disease Detection}
	\label{sec:Pb_fish_detector}
	The symptoms of flatfish diseases can be displayed anywhere on the body. 
	The appearance of disease symptoms is more related to the body part of the fish rather than the types of diseases as depicted in FIGURE~\ref{fig:appearance-bodyparts}. 
	The goal of our study is not to classify the disease types, but to detect the diseases immediately whether any symptoms have been visually exposed on the flatfish.
	To achieve this, we focus only on the visual features of diseases.
	Therefore, this section proposes dividing the flatfish into three different parts (i.e., head, body, fins) to evaluate them separately (FIGURE~\ref{fig:framework}).
	
	\begin{figure}[t]
		\centering
		\subfigure[Detection 1]{\includegraphics[width=0.18\linewidth]{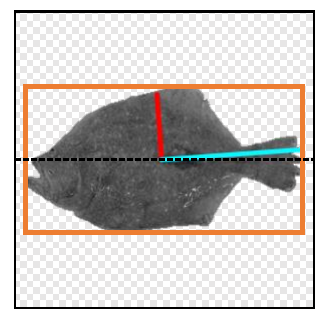}}
		\subfigure[Detection 2]{\includegraphics[width=0.18\linewidth]{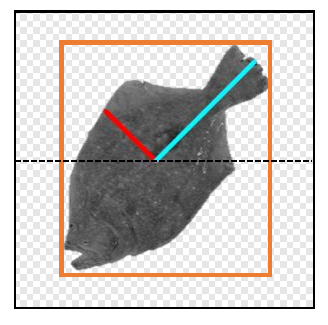}}
		\subfigure[Detection 3]{\includegraphics[width=0.18\linewidth]{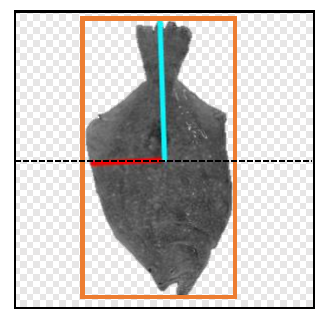}}
		\subfigure[Manual alignment]{\includegraphics[width=0.18\linewidth]{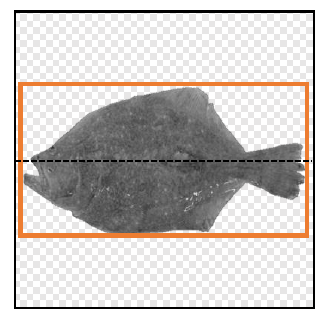}}
		\hspace{5pt}
		\subfigure[Proposed alignment]{\includegraphics[width=0.18\linewidth]{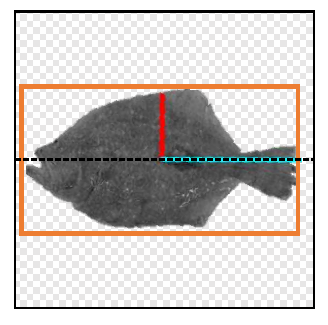}}
		\caption{Examples of the processes for the horizontal fish alignment: (a)--(c) examples of detection results of different fish directions, (d) manual alignment, and (e) aligned results based on the principal component analysis (PCA). Yellow boxes indicate the bounding boxes of the fish. Cyan lines (\textcolor{cyan}{---}) denote the principal axis of $\mathcal{C}_n$. Black dotted lines indicate the horizontal line.}
		\label{fig:horizontalfish}
	\end{figure}
	
	\subsection{Fish Localization and Horizontal Alignment}
	\label{subsec:local_alignment}
	
	An olive flounder detector was trained to localize fish in the images by automatically finding the position and size of the fish in the image. 
	The detection result consists of a four-dimensional vector $\mathbf{D}_{n} = \left(x_{n}, y_{n}, w_{n}, h_{n}\right)$, where $\left(x,y\right)$ is the center position, $\left(w,h\right)$ denotes the width and height of the fish, and $n$ represents the index of the detection, respectively.
	Any object detectors, such as R-CNN~\cite{ren2015faster}, or the YOLO series~\cite{redmon2016youu,ge2021yolox,yolov5} detectors, can be used for fish localization.
	Rectangular bounding boxes~(bbox) $(\mathbf{D}_{n})$ of the detected fish are obtained, but those rough bbox still include many background areas due to tilted fish directions, as illustrated in FIGURE~\ref{fig:horizontalfish}~(a)--(c). A possible solution is to manually align the fish along the horizontal axis when taking images (e.g., FIGURE~\ref{fig:horizontalfish}~(d)). However, a wide variety of fish shapes exist, and aligning live and swimming fish is difficult.
		
	\begin{figure}[t]
		\centering
		\subfigure[]{\includegraphics[width=0.4\linewidth]{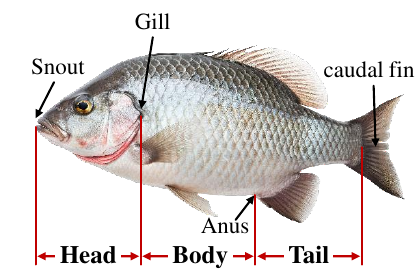}}
		\hspace{10pt}
		\subfigure[]{\includegraphics[width=0.4\linewidth]{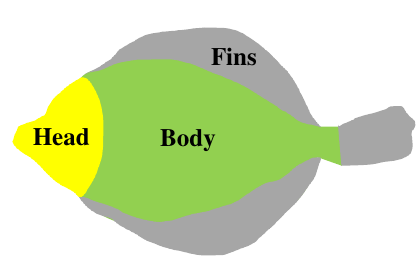}}
		\caption{(a) Conventional category of fish parts in a fishery study~\cite{yu2023comparative}. (b) Category of flatfish for disease detection in this study. We categorize the flatfish into head, fins, and body segments.}
		\label{fig:conventional-fishpart}
	\end{figure}
	
	Therefore, we propose a horizontal alignment method for detected fish.
	Employing image segmentation and principal component analysis (PCA)~\cite{mackiewicz1993principal} methods.
	Given bbox of the fish $\mathbf{D}_{n}$, the corresponding rectangular images as cam be cropped as $\mathbf{I}_{n}$.
	Performing foreground segmentation for the cropped images $(\mathbf{I}_{n})$, yields a set of $(x,y)$ pixel coordinates of the fish foregrounds represented as follows:
	\begin{equation}
		\label{math:1}
		\mathcal{C}_n = \{\left(x_k,y_k\right)|1 \leq k \leq  K_n\},
	\end{equation}
	where $K_n$ is the number of foreground pixels in the $n$th cropped image.
	Fish segmentation was conducted using U2-Net~\cite{Qin_2020_PR}, but any method can be used. 
	The method works unsupervised and does not require training steps for fish segmentation.
	After segmentation, the covariance matrix of $\mathcal{C}$ is calculated as follows:
	\begin{equation}
		\label{math:2}
		\mathbf{\Sigma} = \frac{1}{K}\left(\begin{matrix}
			\sum(x_k-m_x)^2  & \sum(x_k-m_x)(y_k-m_y) \\
			\sum(x_k-m_x)(y_k-m_y) & \sum(y_k-m_y)^2 \\
		\end{matrix}\right),
	\end{equation}
	where $m_x$ and $m_y$ are the means of the $x$ and $y$ values, respectively.
	Based on PCA, two eigenvalues and eigenvectors are calculated from the covariance matrix $\mathbf{\Sigma}$, and the eigenvector $\mathbf{v}_{\lambda}$ corresponding to a larger eigenvalue~$\lambda$ is selected. 
	Then, the eigenvector $\mathbf{v}_{\lambda}$ depicted by the cyan line (\textcolor{cyan}{---}) in FIGURE~\ref{fig:horizontalfish} reflects the rotation of the fish.
	
	By rotating the eigenvector $\mathbf{v}_{\lambda}$ to lie on the horizontal line, the fish are aligned regardless of their initial detection results as in FIGURE~\ref{fig:horizontalfish}(e).
	The proposed fish alignment method works well because the flatfishes almost like a two-dimensional object on the floor.
	Compared to the manual alignment in FIGURE~\ref{fig:horizontalfish}(d), the proposed method performs accurately aligns diverse flatfish horizontally.

	\begin{figure}[t]
		\centering
		\includegraphics[width=0.5\linewidth]{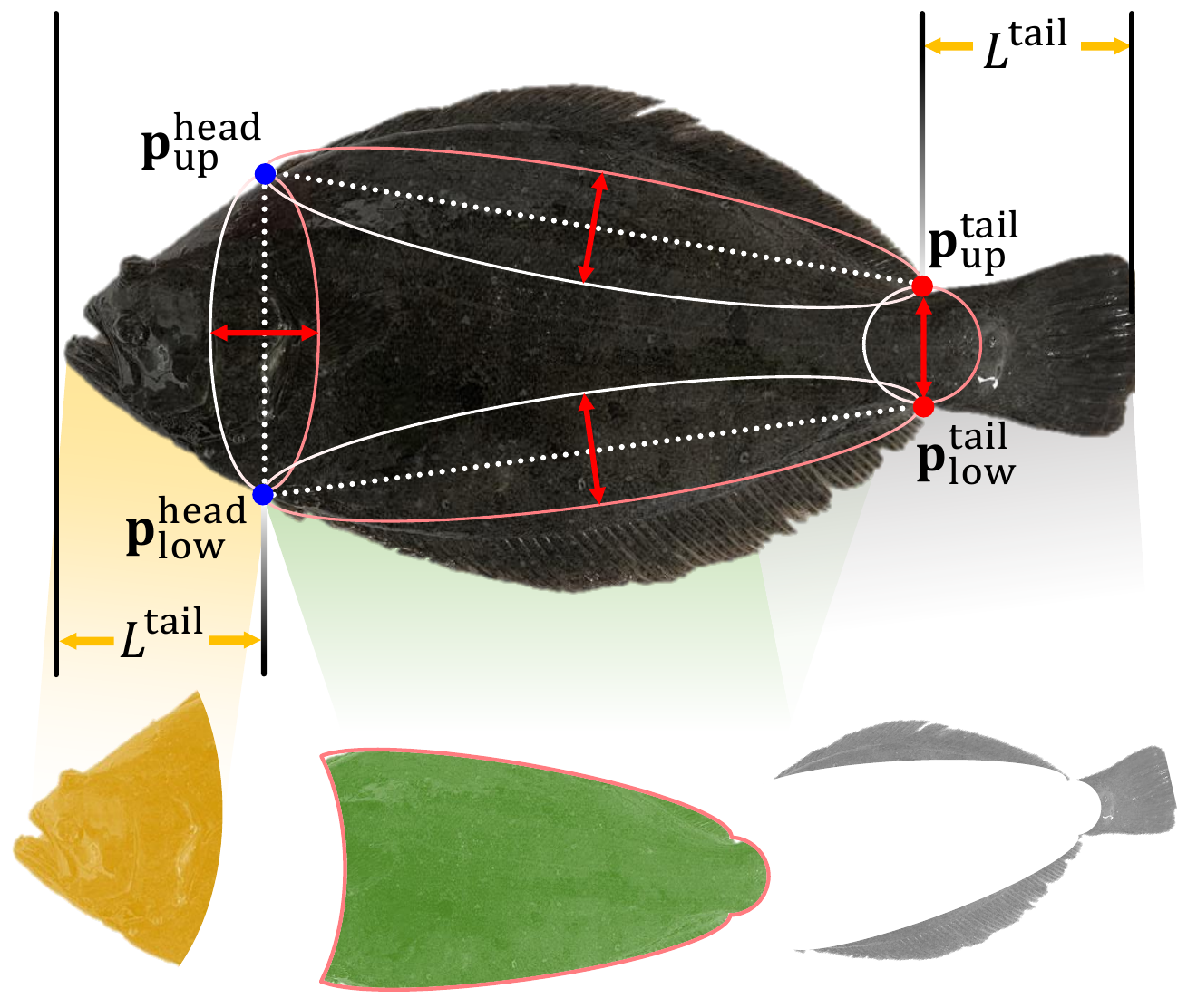}
		\caption{An illustration of the flatfish feature points and three segments. The feature points for the head and tail are denoted by blue $\textcolor{blue}{\bullet}$ and red $\textcolor{red}{\bullet}$ dots. A red (\textcolor{red}{$\leftrightarrow$}) and orange (\textcolor{orange}{$\leftrightarrow$}) arrows denote the thickness and length of the tail, respectively. Given feature points and the lengths of the arrows, three ellipses and one circle are fitted and segment flatfish into distinctive parts.}
		\label{fishpartsegment}
	\end{figure}
	
	\subsection{Fish Part Segmentation for Disease Detection}
	\label{sec:Three-part split}
	
	In the fisher biology field, fish parts are typically categorized into three main sections: head, body, and tail. 
	According to the report~\cite{yu2023comparative}, the head includes from the snout to the gill, the body expends from the gill to the anus, and the tail covers from the anus to the front of the caudal fin, as illustrated in FIGURE\ref{fig:conventional-fishpart}~(a). 
	In this study, we observed that fish parts can be categorized differently according to the types of their diseases.
	Based on a careful examination of the diseases in our dataset, we newly categorized fish parts as follows. 
	\begin{itemize}
		\item[1)] The head: covers the snout, eyes, and a gill. Unlike other fish, both eyes of the flatfish can be examined at the same time since their eyes are exposed on one side.
		\item[2)] The fins: include the upper, and lower fins and a caudal fin, except for the pectoral fin, which is excluded because flatfish pectoral fins are relatively smaller than those of other fish.
		\item[3)] The body: covers the rest of the fish except for the head and fins.
	\end{itemize}
	FIGURE~\ref{fig:conventional-fishpart}~(b) illustrates the three parts of fish for disease detection.
	
	To segment the defined fish parts, we can consider deep learning-based approaches~\cite{yu2018bisenet,strudel2021segmenter,gurita2021image}.
 	However, training deep learning networks for all the proposed frameworks is redundant. 
  	In general, the segmentation method requires a huge amount of training data for network training, but the dataset of flatfish is quite rare. 
	Therefore, this study proposes a fish-part segmentation method that can analyze the shapes of fish and automatically find the three parts without supervised learning stages.
	To achieve this goal, we initially detect the position of tails and measure their pixel length. 
	The tails of flatfish play a crucial role in comprehending their overall structure. Despite variations in individual fish, a proportional and similar relationship exists between the tail and head lengths. 
	Tails are relatively easier to locate compared to other fish parts due to their distinct structural features.
	
	\begin{figure}[t]
		\centering
		\subfigure[Ocular-side (topside)]{\includegraphics[width=0.45\linewidth]{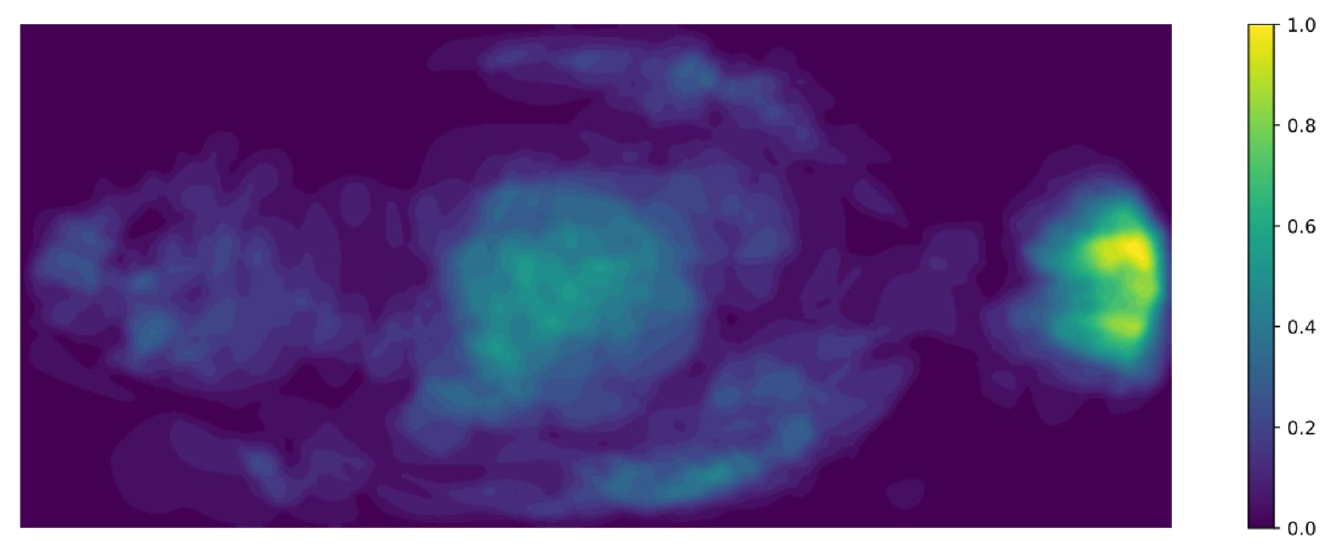}}
		\subfigure[Blind-side (bottom side)]{\includegraphics[width=0.45\linewidth]{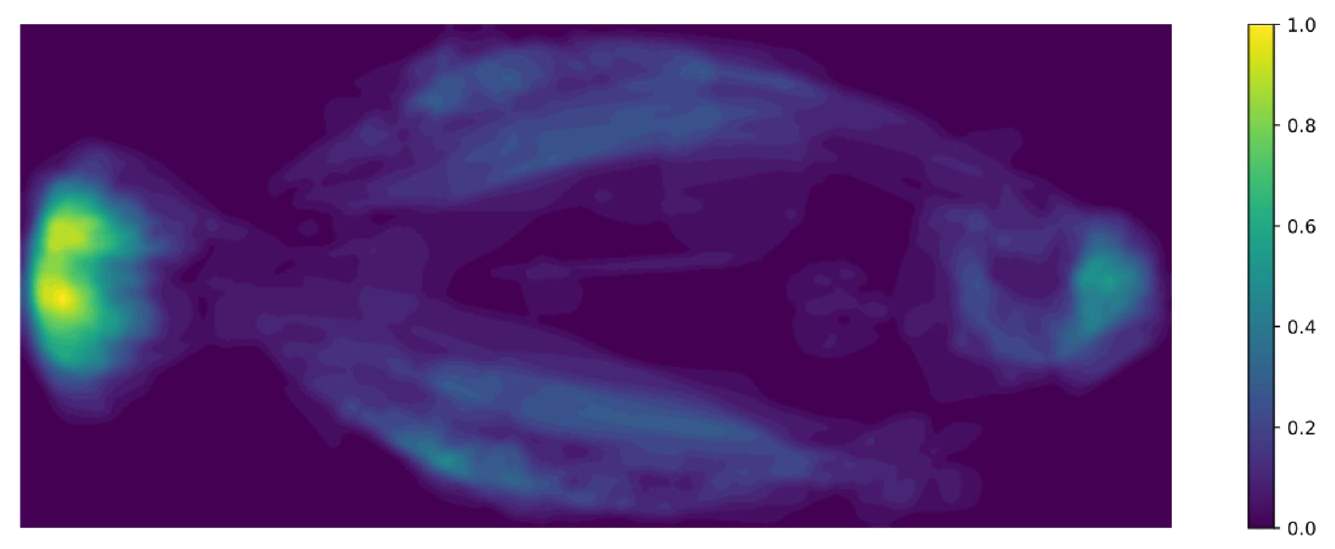}}
		\caption{Disease occurrence heatmaps of the ocular and blind-side in the \texttt{FlatIMG} dataset. The scores of the heatmaps were normalized in $\left[0,1\right]$. (a) On the ocular side, diseases occur primarily around the gills, fins~(especially the tail fins), and centers of their bodies. (b) On the blind-side, diseases also occur around the fins and gills. The occurrence score of mouth on blind-side is higher than the score on the ocular-side. Diseases rarely occur on the body part of the blind-side~(i.e., the stomach area) compared to the ocular-side.}
		\label{fig:data_heat_map}
	\end{figure}
	
	\begin{figure*}[t]
		\centering
		\includegraphics[width=\linewidth]{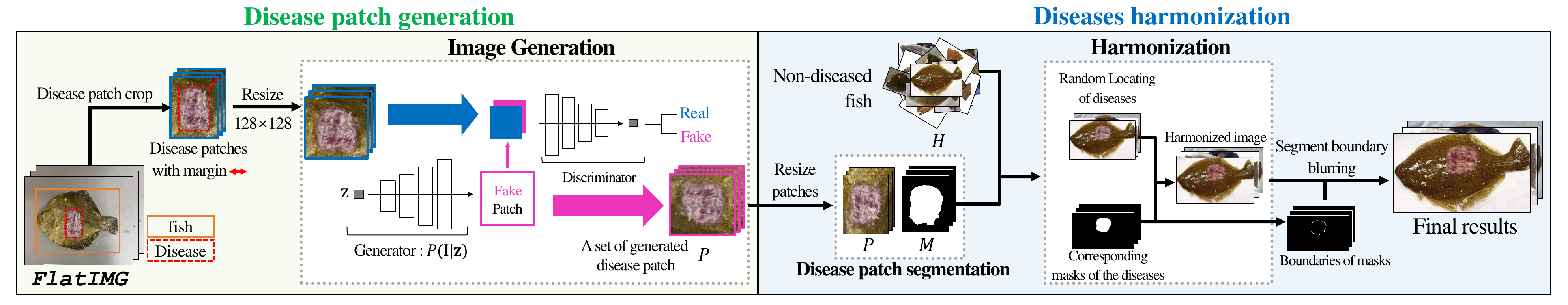}
		\caption{Overall process for diseased fish image generation, consisting of two methods: 1) disease image patch generation; 2) disease harmonization. In the disease patch generation process, given flatfish images~(\texttt{FlatIMG} in Section~\ref{sub:flatIMG}) with diseases are cropped and input into the generative model. The trained model can generate fake but realistic disease patches by manipulating an input vector $\mathbf{z}$. In the disease harmonization process, non-diseased fish images $\mathcal{H}$ and generated disease patches $\mathcal{P}$ are mixed. The location and size of disease patches are randomly determined on the flatfish. For more naturally appearing harmonization, the boundaries of diseases are smoothed.}
		\label{fig:generation-process}
	\end{figure*}
	
	Owing to the horizontal alignment proposed in Section~\ref{subsec:local_alignment}, the tail of the flatfish can be found with the following steps. 
	Initially, a set of fish boundary pixels $b\left({C}_n\right)$ from the fish foreground points ${C}_n$ are determined.
	These boundary pixels $b^{}\left({C}_n\right)$ are then divided into two subsets: the upper and the lower boundaries, as illustrated in FIGURE~\ref{fishpartsegment}.
	For the upper boundary, the local minimum point $\mathbf{p}^{(tail)}_{up}$ is estimated, and for the lower boundary,the local maximum point $\mathbf{p}^{(tail)}_{low}$ is estimated. 
	Then, the center position of the tail can be defined by $\mathbf{p}^{(tail)}_{c} = \frac{\mathbf{p}^{(tail)}_{up}+\mathbf{p}^{(tail)}_{low}}{2}$.
	These points determine the position of the tail and the tail length $L^{(tail)}$ can be measured by calculating the distance between $\mathbf{p}^{(tail)}_{c}$ and the end of the tail position as shown in FIGURE~\ref{fishpartsegment}.
	
	Once the tails are found, the head part can be estimated by fitting an ellipse around the gill of the fish.
	Thus, the center of the head is set to $\mathbf{p}^{(head)}_{c}$ based on the position of the snout and tail length $L^{(tail)}$. Then, the upper and lower positions of the head $\mathbf{p}^{(head)}_{up}, \mathbf{p}^{(head)}_{low}$ can be determined along with the vertical direction from $\mathbf{p}^{(head)}_{c}$.
	Finally, an ellipse can be fit with two axis lengths: the major axis, defined as the line between two head points $\mathbf{p}^{(head)}_{up}$ and $ \mathbf{p}^{(head)}_{low}$, and the minor axis, defined as the tail length $L^{(tail)}$.
	Similarly, the upper and lower fins are defined based on the ellipsoid fitting between the obtained points and tail length.
 	The overall processes are depicted in FIGURE~\ref{fishpartsegment}.
	Based on the ellipses and points, the three parts of the fish can be separated.
	Then, three different disease detectors are trained for these parts. 
	This approach enables effectively evaluating different disease appearances.
	
	\begin{figure}[t]
		\centering
		\includegraphics[width=0.8\linewidth]{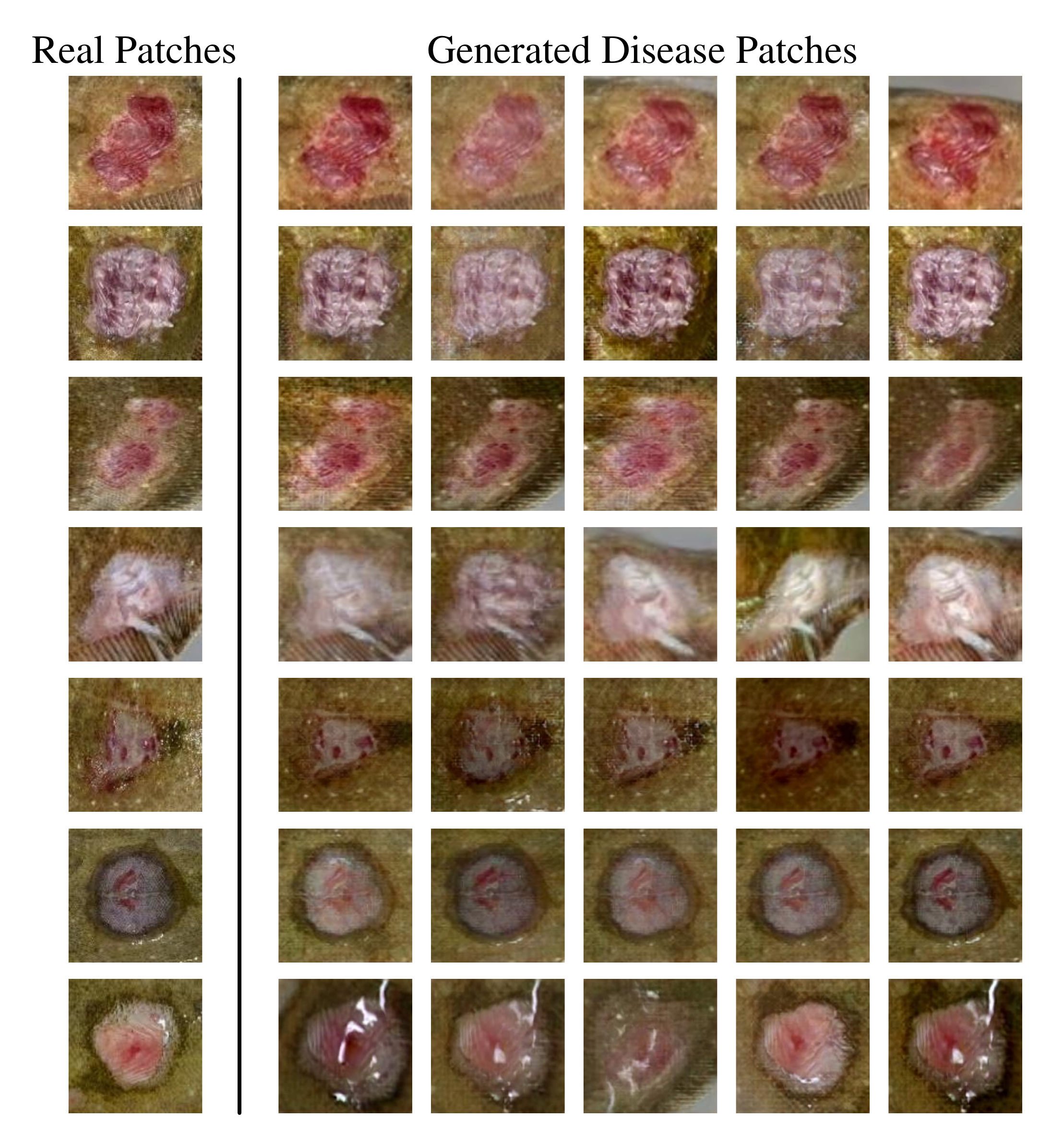}
		\caption{Fake but realistic disease patches generated based on the trained generator $p\left( \mathbf{I} | \mathbf{z} \right)$. Images in the first column are the real disease image patches that were utilized to train the generator. Images from the second to sixth columns are generated disease patches by manipulating the value of the noise vector $\mathbf{z}$.}
		\label{fig:generation-patches}
	\end{figure}
	
	\begin{table}[t]
		\centering
		\caption{Number of diseases by body part and side.}
		\small
		\setlength\tabcolsep{4.5pt}
		\renewcommand{\arraystretch}{1}
		\begin{tabularx}{\columnwidth}{>{\centering\arraybackslash}X||>{\centering\arraybackslash}X|>{\centering\arraybackslash}X|>{\centering\arraybackslash}X}
			\hline\noalign{\hrule height 1pt}
			\rowcolor[HTML]{EFEFEF} 
			Part \textbackslash Side  & Ocular-side & Blind-side & Total  \\ \hline
			Head         & 49         & 51        & 100 \\ \hline
			Fins         & 87         & 91        & 178 \\ \hline
			Body         & 70         & 46        & 116 \\ \hline
			\noalign{\hrule height 1pt}
		\end{tabularx}
		\vspace{-5pt}
		\label{tab:data_body_parts}
		\vspace{-5pt}
	\end{table}
	
\begin{table}[t]
    \centering
\caption{Number of diseases by pathogen type.}
    \begin{threeparttable}
        \renewcommand{\arraystretch}{1}
        \begin{tabular}{r||c|c}
            \hline\noalign{\hrule height 1pt}
            \rowcolor[HTML]{EFEFEF} 
            Pathogen type  & Pathogen name & Occurrence \\ \hline
            \multirow{9}{*}{Bacteria} & \textit{S. parauberis} & 42 \\ 
            \cline{2-3} & \textit{V. scophthalmi} & 13 \\ 
            \cline{2-3} & \textit{V. harveyi} & 6 \\ 
            \cline{2-3} & \textit{V. sinensis} & 2 \\ 
            \cline{2-3} & \textit{V. tapetis} & 2 \\ 
            \cline{2-3} & \textit{V. fischeri} & 2 \\ 
            \cline{2-3} & \textit{E. tarda} & 5 \\ 
            \cline{2-3} & \textit{E. piscicida} & 19 \\ 
            \cline{2-3} & \textit{Photobacterium damselae} & 11 \\ 
            \hline
            \multirow{2}{*}{Viruses} & \textit{VHSV} & 5 \\ 
            \cline{2-3} & \textit{FHM,HINAE} & 3 \\  
            \hline 
            \multirow{3}{*}{Parasites} & \textit{Scuticociliates} & 38 \\ 
            \cline{2-3} & \textit{Trichodinids} & 30 \\  
            \cline{2-3} & \textit{Skin Flukes} & 2 \\  \hline
            \noalign{\hrule height 1pt}
        \end{tabular}
        \begin{tablenotes}[flushright]
            \scriptsize
            \item S.: streptococcus
            \item V.: vibrio
            \item E.: edwardsiella
        \end{tablenotes}
    \end{threeparttable}
    \vspace{10pt}
    \label{tab:typesofdis}
    \vspace{-5pt}
\end{table}

	\section{Fish Disease Image Dataset}
	\label{dataset}
	\subsection{Real Flatfish Disease Images: \texttt{FlatIMG}}
	\label{sub:flatIMG}
	As mentioned in Section~\ref{sec:Related works}, not public image datasets contain fish diseases with annotations.
	Therefore, we newly collected and categorized a fish disease image dataset named \texttt{FlatIMG}.
	We collected $164$ diseased flatfish from $10$ different fisheries in Jeollanam-do, South Korea.
	Among the $164$ fish identities, the number of visually identified diseases is $394$.
	Each fish was photographed on both the close-up ocular-side~(top side) and blind-side(bottom side) with high-resolution digital cameras (2048$\times$1536).
 	As stated in Section.~\ref{sec:introduction}, the goal of our study is not to classify the disease types or pathogens, but to detect the diseases immediately whether any symptoms have been visually exposed on the fish.
	Therefore, we set the three main areas of the fish (Head, Fins, and Body), as proposed in Section~\ref{sec:Three-part split}, and categorized \texttt{FlatIMG} by body part and side in Table.~\ref{tab:data_body_parts}.
	There are $100$ diseases that occur on the head, $178$ on the fins, and $116$ on the body parts, respectively.
	Some of the diseases were photographed multiple times with diverse camera angles to carefully examine the diseases.
	To summarize, the total number of disease images is $1,861$ (including images photographed multiple times) and disease identities are $394$ in \texttt{FlatIMG}.

	In Table.~\ref{tab:typesofdis}, the types of pathogen-occurring diseases are categorized by bacteria, virus, or parasite infections. 
	In bacterial infections, nine pathogens were recorded: \textit{Streptococcus parauberis}, \textit{Vibrio scophthalmi}, \textit{V.~harveyi}, \textit{V.~sinensis}, \textit{V.~tapetis}, \textit{V.~fischeri}, \textit{Edwardsiella tarda}, \textit{E.~piscicida}, and \textit{Photobacterium damselae}. 
	Virus infections include two pathogens: such as \textit{VHSV} (viral hemorrhagic septicemia virus)  and two types of rhabdovirus (\textit{FHM} and \textit{HINAE} ).
	Parasite infection includes three pathogens: \textit{scuticociliates}, \textit{trichodinids}, and \textit{skin flukes}.

   	While diseases and pathogens do not have a one-to-one mapping, we included detailed annotations on various diseases that can be caused by different pathogens, which helps understand the diverse symptoms associated with each pathogen, enabling more accurate diagnoses and research. 
   	In future studies, this dataset can be utilized to develop disease detection models, contributing to the early diagnosis and prevention of various diseases.

	\begin{figure*}[t]
		\centering
		\includegraphics[width=0.85\linewidth]{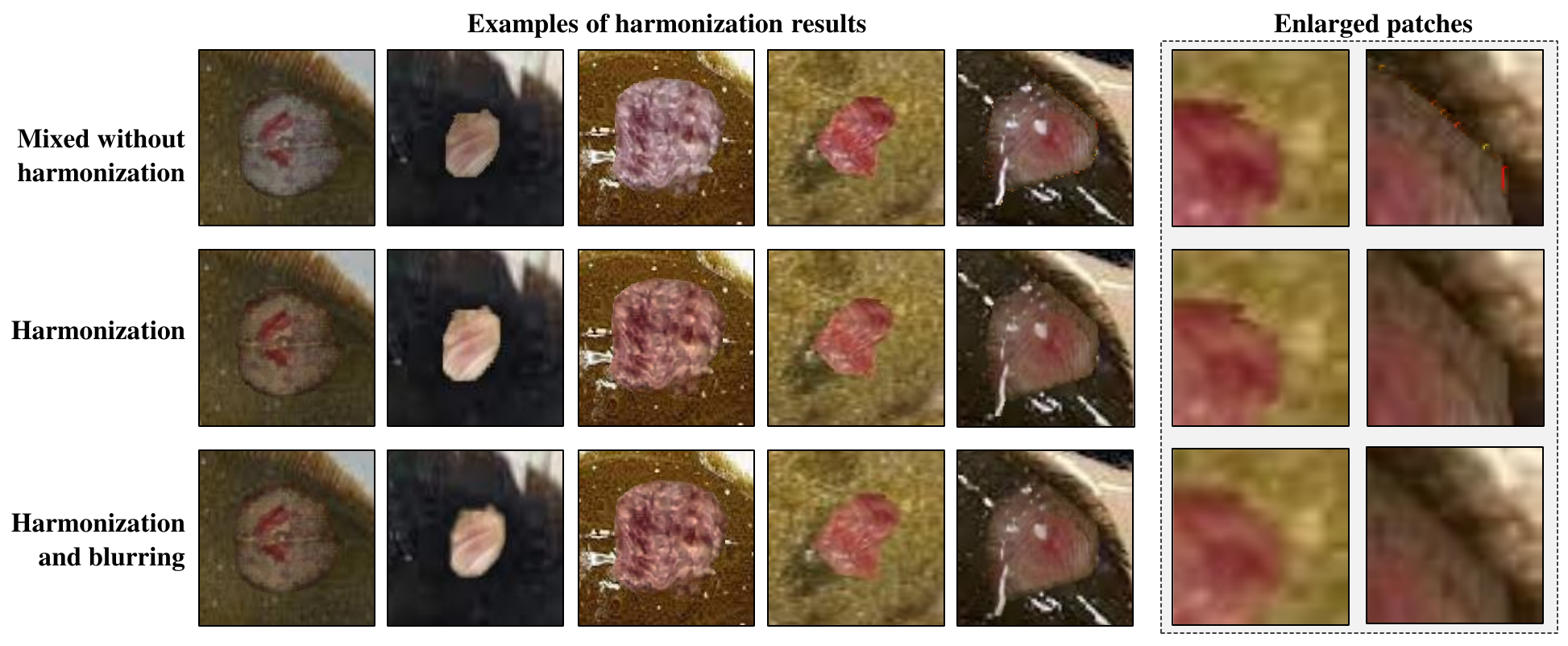}
		\caption{Examples of image harmonization results between generated disease patches and non-diseased flatfish. First row: Mixed images without an image harmonization step. 
			The colors of the mixed areas are unnatural to their surrounding areas.
			Second row: Harmonization results, harmonizing well with the color distributions of the target images. The boundaries of source areas are still unnatural due to the large pixel value differences around the boundaries.
			Third row: Final results of harmonization including boundary blurring, resulting in very natural-looking diseased flatfish images.}
		\label{fig:harmonizationresults}
	\end{figure*}
	
	Several diseased flatfish images in \texttt{FlatIMG} are shown in FIGURE~\ref{fig:appearance-bodyparts}.
	The disease shapes are diverse and can occur anywhere on the fish.
	In addition, the heatmaps of the diseases are plotted in FIGURE~\ref{fig:data_heat_map}.
	They show where the diseases occur on the ocular-side and blind-sides of the flatfish at a glance. Based on the heatmaps, we noted that diseases mainly occur around gills and fins on both sides.
	These diseases rarely occur on the body of the blind-side~(i.e., the stomach) compared to ocular-side.
	
	To the best of our knowledge, this is the largest public fish disease image dataset. 
	However, both the amount of the dataset and the identities of fishes are still insufficient to validate the proposed methods without bias and over-fitting problems. 
	To overcome the lack of the dataset, many methods have performed image augmentations that augment the original images based on color and geometric transformations~\cite{maharana2022revieww}.
	However, simple color and geometric transformations rely on the original distribution of the reference image. 
	Thus, the variations of the augmented images are limited.
	Although these approaches work well for common objects, when it comes to fish disease generation, the augmented result looks unnatural and artificial.
	The following section proposes a fish disease image-generation framework based on the generative adversarial networks~(GAN) and image harmonization methods. 
	
	\subsection{Flatfish Disease Image Generation}
	\label{sub:dataset_gen}
	
	The proposed augmentation framework aimed to enrich the dataset by generating natural-looking fish images with diseases that closely mimic real data.
	The framework consists of two main methods: 1) disease image patch generation; and 2) disease harmonization.
	The overall framework is described in FIGURE~\ref{fig:generation-process}.
	
	\subsubsection{Disease Patch Generation}
	For the training data, the diseased patches were cropped with a sufficient margin to include the whole boundaries of the disease pattern.
	Then, an image generator was trained using distributions of the disease patches.
	The generator is represented by $p\left( \mathbf{I} | \mathbf{z} \right)$, where the conditional variable $\mathbf{z}$ denotes an input noise vector, and $p\left( \mathbf{I} | \mathbf{z} \right)$ denotes the generated distribution of the disease image given $\mathbf{z}$.
	Based on the generator, $p\left( \mathbf{I} | \mathbf{z} \right)$, new disease patches can be created by manipulating the input noise vectors $\mathbf{z}$ as shown in FIGURE~\ref{fig:generation-patches}.
	As a result, a set of generated disease image patches is denoted by
	\begin{equation}
		\mathcal{P} = \{p\left( \mathbf{I}_i | \mathbf{z}_i \right)|1 \leq i \leq  N\}, 
	\end{equation}
	where $N$ is the total number of generated images.
	The cardinality of the set is $\left|\mathcal{P}\right|= N$, and the patches represent various square disease patterns.
	Not all areas of the generated patch are diseases, because they contain healthy areas around the diseases.
	The healthy area is called the `margin' in this study.
	It was confirmed that some of the margin surrounding diseases significantly help to train a reliable generator.
	Due to the margin area, the generator can learn patterns of disease boundaries.
	
	\subsubsection{Disease Harmonization}
	Given the generated disease patches, $\mathcal{P}$, synthetic flatfish images were generated with various diseases.
	Thus, several flatfish images were prepared with no diseases, denoted by $\mathcal{H}$, and can perform image harmonization that naturally mixes two different image sources: $\mathcal{P}$ and $\mathcal{H}$.
	The harmonizing process requires foreground masks of the harmonizing patches $\mathcal{P}$ to naturally blend the foreground areas~\cite{ke2022harmonizer}.
	Therefore, foreground disease areas of $\mathcal{P}$ were estimated based on segmentation methods in Section~\ref{subsec:local_alignment}. 
	A set of foreground segment masks is denoted by $\mathcal{M}$, corresponding to $\mathcal{P}$.
	The generated disease patches $\mathcal{P}$ with their foreground masks $\mathcal{M}$ were set as the source images, and the non-diseased flatfish images $\mathcal{H}$ were set as the target images.
	Finally, image harmonization of source images is performed onto target images.
	
	\begin{figure}[t]
		\centering
		\includegraphics[width=0.8\linewidth]{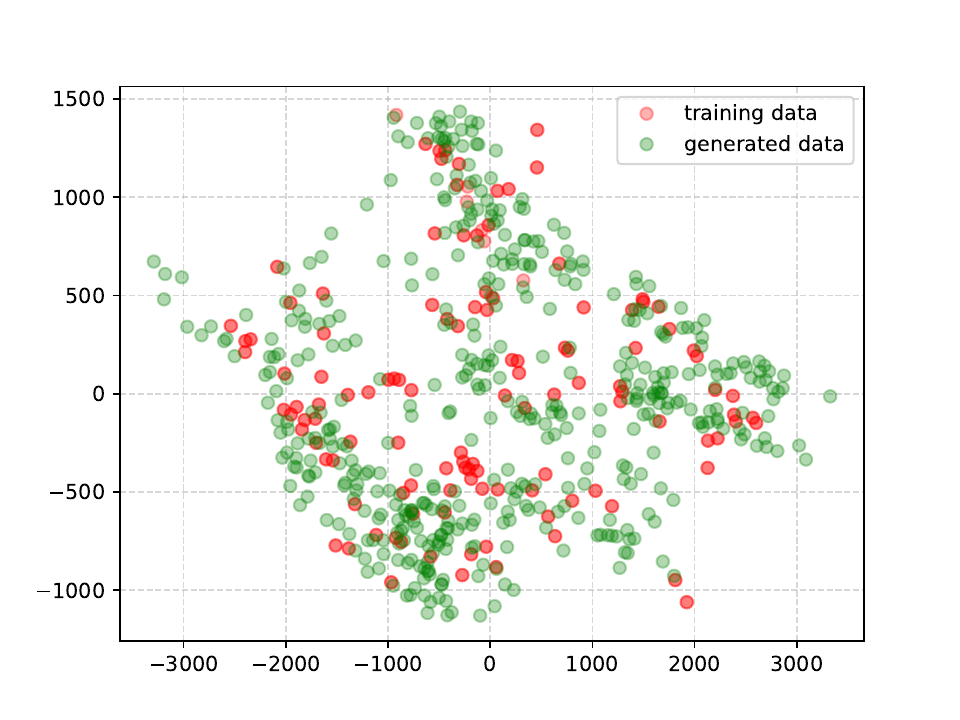}
		\caption{Distributions of training disease patches and generated disease patches by the generative model. The generated patches are similar to the training data, but not the same with them.}
		\label{fig:distributions_train-gener}
	\end{figure}

 	\begin{figure*}[t]
		\centering
		\includegraphics[width=1\linewidth]{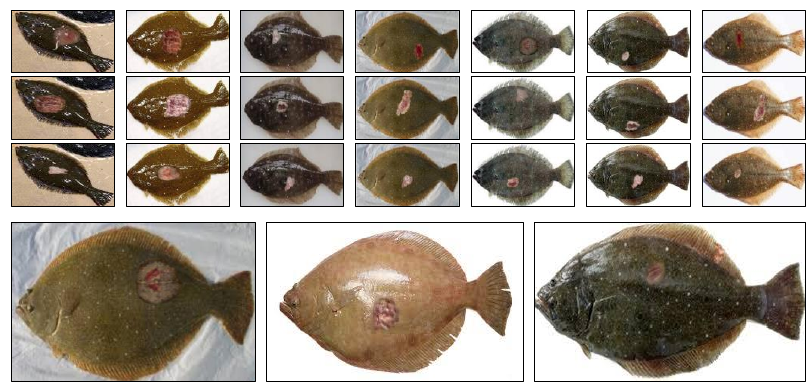}
		\caption{Examples of generated diseased flatfish images. The harmonizer mixes $462$ disease patches generated by the trained model $p\left( \mathbf{I} | \mathbf{z} \right)$ and $59$ non-disease flatfish images. The overall quality of the generated images is natural. The total number of the generated images is $4,773$.}
		\label{fig:generationresults}
	\end{figure*}
	
	FIGURE~\ref{fig:harmonizationresults} illustrates the results of image harmonization. 
	The first row shows the mixed images without image harmonization.
	They simply replaced the area of the target image with the values of the foreground patch.
	The color of the mixed area is quite unnatural with its surrounding areas.
	In contrast, the mixed results based on image harmonization are depicted in the second row of FIGURE~\ref{fig:harmonizationresults}, and harmonize well with the color distributions of the target images. However, the boundaries of the source areas are still unnatural due to the large pixel value differences around the boundaries.
	To handle this problem, we further perform blurring the boundary of foreground patches to make more naturally harmonized images.
	The third row of FIGURE~\ref{fig:harmonizationresults} shows the final results of the harmonization including boundary blurring.
	As in the figure, the unnaturalness of disease boundaries is mitigated resulting in natural-looking diseased flatfish images.
	
	An advantage of the proposed framework is that it produces not only natural but also a wide variety of image-generation results.
	For example, there are $\mathcal{\left|P\right|} \times \mathcal{\left|H\right|}$ possible image harmonization combinations, the size and position of the source images $\mathcal{P}$ can be manipulated on the target images $\mathcal{H}$.
	The variations of source images are denoted by $N_s$; and the possible number of combinations for image harmonization is  $\mathcal{\left|P\right|} \times \mathcal{\left|H\right|} \times N_s$.
	All code for the proposed disease image generation framework, the datasets including \texttt{FlatIMG}, and generated images are available at \url{https://will_be_available}.
	\section{Experimental Results}
	\label{sec:Experimental Results}
	
	\subsection{Experimental Settings and Evaluation Metrics}
	To validate the proposed methods, evaluated two fish disease image datasets, the--\texttt{FlatIMG} dataset and a salmon image dataset that we named \texttt{SalmIMG}~\cite{ahmed2024salmonscan}.
	The details for the \texttt{FlatIMG} dataset are explained in Section~\ref{sub:flatIMG}.
	Like flatfish, salmon is well known to be highly farmed and consumed worldwide. 
	Therefore, we also tested salmon for the disease detection task.
	\texttt{SalmIMG} dataset includes 82 images of salmon with visible diseases. 
	However, the dataset does not provide the ground-truth for disease.
	Thus, the ground-truth for the disease regions and classes were manually annotated.          	
	For an unbiased evaluation, five-fold cross-validation was performed in all experiments.
	We utilized You Only Look Once Version 8 (YOLOv8)~\cite{url:yolov8} as the baseline disease detector.
	The YOLOv8m model was trained using NVIDIA RTX 3060ti with the following hyper-parameter settings: 200 epochs, a batch size of 4, and a learning rate of 0.0001.
	
	To measure performances the intersection over union~($IoU$) between two bounding boxes is calculated as $IoU = \frac{\mathbf{D}\cap\mathbf{bbox}_{gt}}{\mathbf{D}\cup\mathbf{bbox}_{gt}}$, where $\mathbf{D}$ is a detection response and $\mathbf{bbox}_{gt}$ represents a ground-truth bounding box of the disease. 
	In the object detection task, a detection response is a true positive~($TP$) where $IoU>0.5$ between $\mathbf{D}$ and $\mathbf{bbox}_{gt}$. 
	If not a matched $\mathbf{bbox}_{gt}$ exists with a detection response, it is a false positive~($FP$).
	Otherwise, if not a matched detection response occurs with $\mathbf{bbox}_{gt}$, it is a false negative~($FN$).
	Then, the recall and precision are calculated as $\frac{TP}{TP+FN}$ and $\frac{TP}{TP+FP}$, respectively.
	Calculating the area under the curve of the precision-recall curve yields the average precision~(AP) score. Finally, the mean average precision~(mAP) is obtained by considering multiple classes of fish diseases.
	The mAP score is a standard evaluation measurement in the object detection research field.

   		\begin{figure*}[t]
			\centering
			\subfigure[Flatfish]{\includegraphics[height=0.3\linewidth]{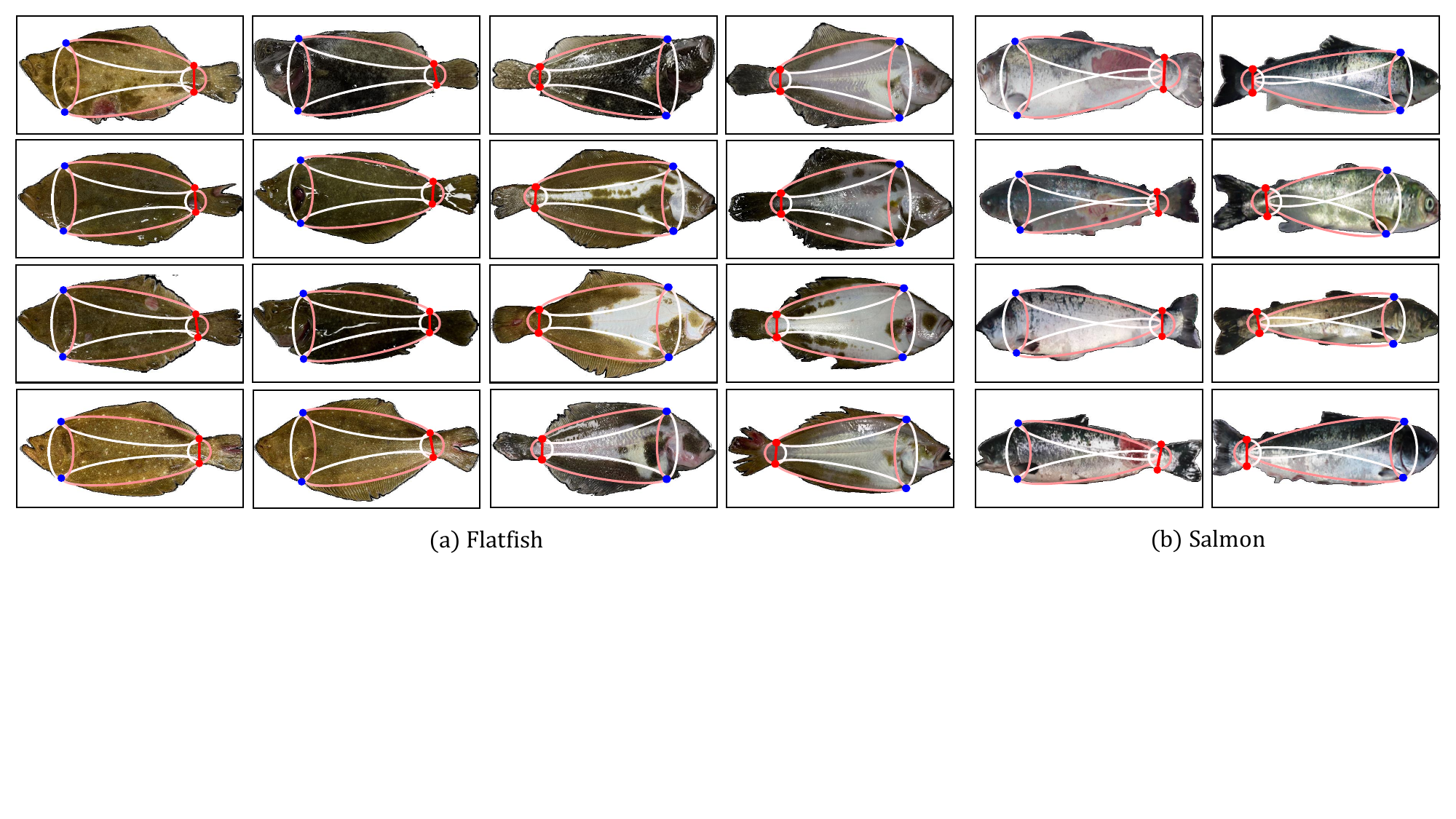}}\hspace{20pt}
			\subfigure[Salmon]{\includegraphics[height=0.299\linewidth]{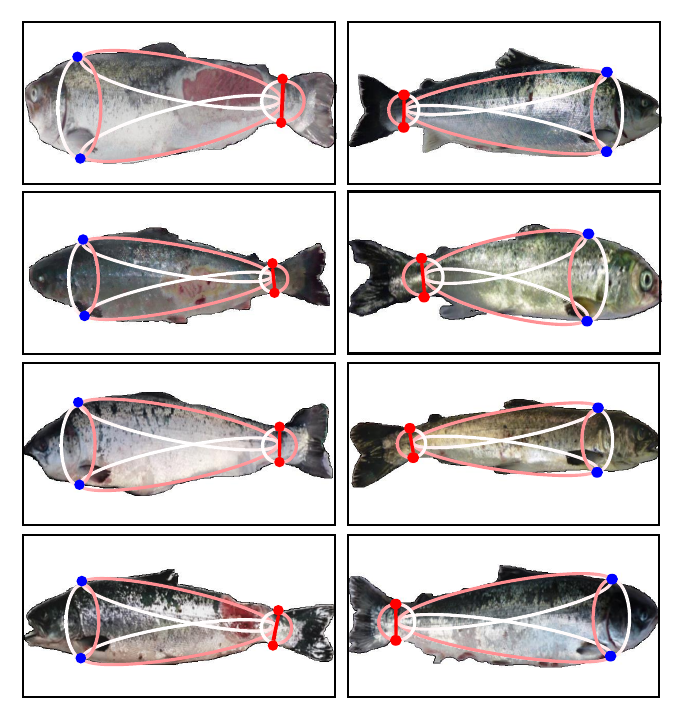}}
			\caption{{Results of applying proposed fish part segmentation to \texttt{FlatIMG} and \texttt{SalmIMG} datasets.}
            The red(\textcolor{red}{―}) line represents the thickness of the tail.
            Based on the length of the thickness of the tail, three ellipses and one circle are fitted to segment the flatfish.}
			\label{fig:3partResult} 
		\end{figure*}
 
	\subsection{Diseased Flatfish Generation Results}

	Examples of generated diseased flatfish images are shown in FIGURE~\ref{fig:generationresults}, and $4,773$ different images were generated with $462$ generated fish diseases and $59$ non-diseased fishes.
	The overall quality of the generated images is natural.
	Note that the proposed framework can perform more diverse diseased fish image generation when it feeds additional source disease and target images.
	Compared to the flatfish dataset, the dataset was augmented to about $9.3$ times the size using the generated images.
	We expect that we can fairly validate the proposed disease detection methods in the following section by using the augmented flatfish images.
	
	Among the real fish disease images collected in Section~\ref{sub:flatIMG}, $240$ body disease patches were selected from $14$ identities, excluding tiny and featureless diseases. 
	That is because, diseases smaller than 100$\times$100 pixel sizes are not enough to learn their complicated patterns.
	In addition, the number of diseases on the body is relatively insufficient compared to other parts, as summarized in Table.~\ref{tab:data_body_parts}.
	An image generator was trained using the distributions of the diseased patches, using StyleGAN2-ADA~\cite{karras2020training} for the generator in the experiment.
	Based on the trained StyleGAN2-ADA model, $462$ new disease patches~(i.e., source images) were generated with the various input vector $\mathbf{z}$.
	
	FIGURE~\ref{fig:distributions_train-gener} shows the distributions of the training patches and the generated patches using the trained model.
	The generated disease patches are similar to the training patches but not identical.
	This result supports that we can generate lots of new disease samples from the model and build new diseased flatfish images.
	For the target images, $59$ disease-free flatfish images were collected from the web.
	Then, the source and target image pairs were randomly matched, setting random locations and sizes for the target fish images.
	For example, the location of the diseased patch should not escape the foreground area of the fish.
	The size of the disease patches is determined within a reasonable range of 80\%--120\% compared to its original size.
	
	In the proposed image harmonization framework, segment anything~\cite{kirillov2023segmentt} was used for disease patch segmentation because it is a zero-shot model that does not require any supervised network refinement steps. 
	In addition, it has been trained by large-scale~($\approx110,000$) and high resolution~($3300\times3950$) image sets, so we expect that it performs superior segmentation accuracy for sophisticated disease textures of flatfish.
	For image harmonization, the method proposed by \cite{ke2022harmonizer}, which can keep the details of input images in terms of textures and styles, was utilized.
	Furthermore, the harmonization method is unaffected by the resolution of the input sources and target images.
	To smooth the boundaries of the segments for natural image harmonization, we performed Gaussian filtering with the $5\times5$ filter of $\sigma = 0.5$.
	
		\subsection{Fish Part Segmentation}
	A method for segmenting fish into three regions (head, fins, and body) was proposed to enhance the detector performance (Section~\ref{sec:Three-part split}).
	The proposed method takes fish images with backgrounds as input, removes the background, and segments the images into three parts. 
	FIGURE~\ref{fig:3partResult}. illustrates the part segmentation results for the fish.
	Despite shape variations among individual fish, we properly detected feature points of fish heads and tails ($\mathbf{p}^{(head)}_{up}, \mathbf{p}^{(head)}_{low}, \mathbf{p}^{(tail)}_{up}, \mathbf{p}^{(tail)}_{low}$) and can divide the fish into three parts.
	Based on the feature points, the ellipses were fitted.
		
		The proposed method was also adopted for the flash salmon disease detection task. 
        Salmon have a thinner body than flatfish; thus, the ellipses were determined slightly differently.
        The ellipses of salmon were generated by connecting a mean-point between $\mathbf{p}^{(tail)}_{up}$ and $\mathbf{p}^{(tail)}_{low}$, and between $\mathbf{p}^{(head)}_{up}$ and $\mathbf{p}^{(head)}_{low}$.
		FIGURE~\ref{fig:3partResult} (b) shows several examples of fish-part segmentation of salmon.
		These experimental results demonstrate that the proposed technique is not only applicable to flatfish but can also be sufficiently applied to other types of fish.
	
        \begin{table}[t]
            \centering
            \caption{Disease detection performance comparison on the \texttt{FlatIMG} dataset. The best and second-best scores are marked in \textcolor{red}{red} and \textbf{bold}, respectively.}
            \small
            \setlength\tabcolsep{4.5pt}
            \renewcommand{\arraystretch}{1}
            \begin{tabular}{l|c}
                \hline\noalign{\hrule height 1pt}
                \rowcolor[HTML]{EFEFEF} 
                Methods  & mean Average Precision (mAP$@$0.5) \\ \hline
                SSD~\cite{liu2016ssd} & 0.055            \\ \hline
                Faster-RCNN~\cite{ren2015faster} & 0.363            \\ \hline
                YOLOv5~\cite{yolov5} & 0.3878           \\ \hline
                YOLOv8~\cite{url:yolov8} & 0.4158            \\ \hline
                Ours -- DA   & 0.3717     \\ \hline
                Ours -- PartSeg  & \textbf{0.4878}     \\ \hline
                Ours -- DA + PartSeg & \textbf{\textcolor{red}{0.5304}} \\ 
                \noalign{\hrule height 1pt}
            \end{tabular}
            \vspace{5pt}
            \label{tab:mAP_flatfish}
            \vspace{-5pt}
        \end{table}
	
        \begin{table}[t]
            \centering
            \caption{Disease detection performance comparison on the \texttt{SalmIMG}~\cite{ahmed2024salmonscan} dataset. The best and second-best scores are marked in \textcolor{red}{red} and \textbf{bold}, respectively.}
            \small
            \setlength\tabcolsep{4.5pt}
            \renewcommand{\arraystretch}{1}
            \begin{tabular}{l|c}
                \hline\noalign{\hrule height 1pt}
                \rowcolor[HTML]{EFEFEF} 
                Methods  & mean Average Precision (mAP$@$0.5) \\ \hline
                SSD~\cite{liu2016ssd} & 0.0298            \\ \hline
                Faster-RCNN~\cite{ren2015faster} & 0.3528            \\ \hline
                YOLOv5~\cite{yolov5} & 0.4328            \\ \hline
                YOLOv8~\cite{url:yolov8} & 0.4374            \\ \hline
                Ours -- DA   & 0.4122     \\ \hline
                Ours -- PartSeg  & \textbf{0.4408}     \\ \hline
                Ours -- DA + PartSeg & \textbf{\textcolor{red}{0.5906}} \\ 
                \noalign{\hrule height 1pt}
            \end{tabular}
            \vspace{5pt}
            \label{tab:mAP_salmon}
            \vspace{-5pt}
        \end{table}

	\begin{figure*}[t]
		\centering
		\includegraphics[width=1\linewidth]{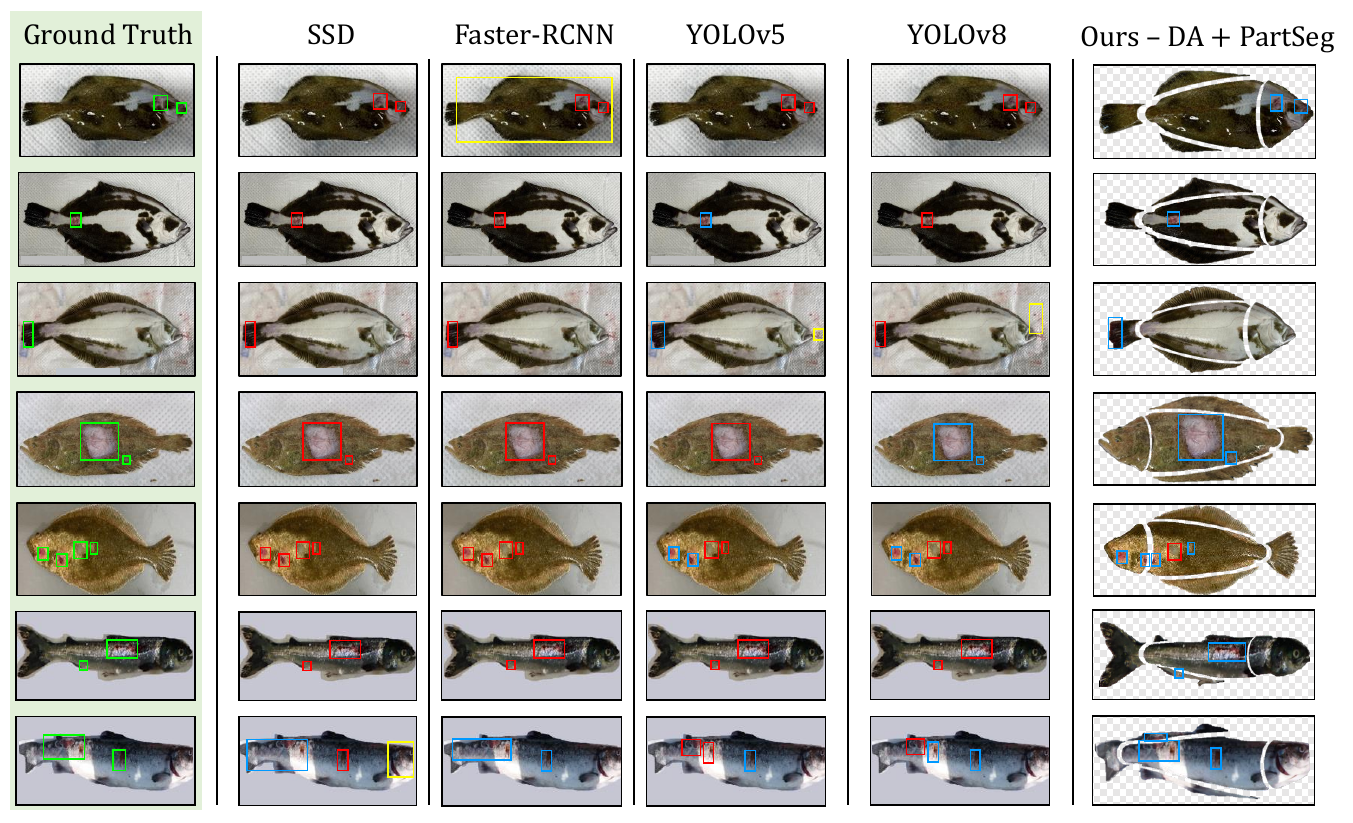}
	\caption{Qualitative results of disease detection. Green box (\textcolor{green}{$\square$}) denotes ground-truth, blue box (\textcolor{cyan}{$\square$}) is true positive~($TP$), red box (\textcolor{red}{$\square$}) is false negative~($FN$), and yellow box (\textcolor{yellow}{$\square$}) is false positive~($FP$). State of the art detection models including SSD~\cite{liu2016ssd}, Faster-RCNN~\cite{ren2015faster}, YOLOv5~\cite{yolov5}, YOLOv8~\cite{url:yolov8} were compared. The YOLOv8~(baseline) shows relatively high detection performance, but Ours-DA+PartSeg shows the best detection results.}
		\label{fig:detectionResults}
	\end{figure*}

 	\subsection{Fish Disease Detection Performance}
		
	    We performed three different scenarios of the proposed methods for disease detection as follows:
	    \begin{itemize}
	    	\item[$\bullet$] 
	    	DA: using augmented fish-disease images proposed in Section.~\ref{sub:dataset_gen} for training fish disease detectors.
	    	\item[$\bullet$] 
	    	PartSeg: performing fish part segmentation proposed in Section.~\ref{sec:Pb_fish_detector}.
	    	\item[$\bullet$] 
	    	DA + PartSeg: using both proposed methods DA and PartSeg for fish disease detection.
	    \end{itemize}
	    Note that, we fixed the number of test images before and after data augmentation~(DA) for a fair comparison.
	
        Table.~\ref{tab:mAP_flatfish} presents a performance comparison in flatfish~(\texttt{flatIMG}) dataset.
        Various object detection models, such as the single-shot detector (SSD)~\cite{liu2016ssd}, Faster-RCNN~\cite{ren2015faster}, YOLOv5~\cite{yolov5}, and YOLOv8 \cite{url:yolov8} were evaluated, with disease detection performance values of 0.055, 0.363, 0.3878, and 0.4158 respectively.
        The SSD failed to properly train disease detection for both fish datasets.
        In contrast, the other detectors such as Faster-RCNN, YOLOv5, and YOLOv8 showed relatively more accurate detection results.
        In addition, YOLOv8, which is the baseline model of the our framework, showed the highest performance among the tested baselines.
        
        In contrast, in our method DA, where image augmentation was applied, the performance decreased to 0.3717. 
        The model generalization performance decreased because it adapted to the augmented data (DA) with increased instances of body diseases.
        Our method using part segmentation~(PartSeg) demonstrates an mAP of 0.4878, marking a 7\% performance increase compared to the baseline method. 
        This result indicates the importance of accurately segmenting fish parts to enhance detection performance.

        Ours--PartSeg can distinguish between different parts of the fish and accurately identify diseases occurring in each part. 
        The generated images provide diversity in the form of disease while preserving the shape and characteristics of the actual fish body parts, enabling more effective learning by the model. 
        Therefore, ours using both proposed methods (DA + PartSeg) exhibits particularly improved performance in detecting diseases in the body area.                
        Overall, these results emphasize the importance of class representation and DA techniques in enhancing the performance of flatfish disease detection models.

        Table.~\ref{tab:mAP_salmon} presents a performance comparison in salmon~(\texttt{SalmIMG}) dataset.
        The same experimental settings and scenarios as in \texttt{FlatIMG} experiments were employed.
        Similar to the results of the flatfish dataset, the performance improved by over 15\% compared to the method of detection as a single class. 
        Thus, the proposed method can be applied not only to flatfish but also to different fish species.
		
	    FIGURE~\ref{fig:detectionResults} illustrates the detection results of various methods.
        The SSD almost failed to detect any diseases.
        The Faster-RCNN showed better detection performance than the SSD, but it often missed the diseases.
        The YOLOv5 exhibits higher detection performance, but it showed several false positives in the background, and could not handle various appearances of diseases.
        The YOLOv8 may seem to have a similar performance to YOLOv5, but looking at the fourth and last detection results, it can be seen that it detected various forms of diseases.
        Our proposed method (Ours - DA + PartSeg) improved the performance of its baseline~(YOLOv8) and robustly detected all diseases that other detectors missed. 
        In particular, it reduced the false positives in the background and detected various types of disease appearances.

	\section{Discussions and Future Work}
	\label{sec:dis_future}
 
	The proposed methods focus on the detection problem rather than disease classification, emphasizing the localization task of finding the lesion location. 
	Therefore, its performance may be somewhat lower compared to simple classification problems. 
	To improve detector performance, securing a large amount of high-quality datasets is generally critical. 
	If diverse lesion data for flatfish is widely available, the detector can better respond to various lesion forms.
	Early detection of fish diseases in aquaculture is a crucial task for preventing infection and improving productivity. 
	While early detection can minimize potential losses, there have been practical infrastructure limitations due to reliance on human labor. 
	By utilizing the method proposed in this study, automatic monitoring of disease lesions can be achieved in actual aquaculture environments.
	Furthermore, we plan to release a high-quality flatfish disease dataset and synthetic images of disease lesions. 
	This will enable widespread use in related research fields, leading to various follow-up studies. 
	Ultimately, our study is expected to contribute to the transition from a human-centered farming environment to a more technology-centered farming environment.

\section{Conclusions}
\label{sec:conclusions}

        Flatfish is one of the major farmed species that is actively cultivated globally and consumed in large quantities. However, the high-density flatfish farming environment makes the fish vulnerable to injuries and diseases, making early disease detection crucial. Traditionally, diseases were detected by human visual inspection, but it was difficult to observe large numbers of fish.
        To overcome these limitations, automated approaches utilizing computer vision and deep learning technologies have been proposed. However, accurate detection remained a challenge as diseases often have diverse forms and small sizes. We utilized Generative Adversarial Networks (GANs) and image harmonization to augment the existing dataset. Through experiments, we confirmed that the augmented dataset was similar to actual lesions.
        Additionally, the fish body was divided into three regions (e.g., head, fins, and body), and the detector was trained for each region. As a result, we achieved 12\% higher performance compared to our baseline model. To verify the generalizability of our proposed method, we applied it to a salmon dataset and achieved a 15\% performance improvement. Through this, we confirmed that our methodology plays a crucial role in enhancing the performance of fish disease detection regardless of the fish species.
	\section*{Acknowledgments}
	
	This research was funded by the Korea Institute of Marine Science Technology Promotion (KIMST), and the Ministry of Oceans and Fisheries, Korea (20220596, Development of Digital the Flow-through Aquaculture System).


\end{document}